\documentclass[10pt]{article}

\usepackage[preprint]{icmlarxiv} 

\usepackage[utf8]{inputenc}
\usepackage[T1]{fontenc}

\usepackage{amsmath,amsfonts,amssymb}
\usepackage{graphicx}
\usepackage{booktabs}
\usepackage{multirow}
\usepackage{xcolor}
\usepackage{enumitem}

\usepackage{hyperref}

\setlist[itemize]{leftmargin=*, topsep=2pt, itemsep=2pt, parsep=1pt}
\setlist[enumerate]{leftmargin=*, topsep=2pt, itemsep=2pt, parsep=1pt}

\begin{document}

\twocolumn[
\icmltitle{When Fine-Tuning Changes the Evidence: Architecture-Dependent Semantic Drift in Chest X-Ray Explanations}

\begin{icmlauthorlist}
  \icmlauthor{Kabilan Elangovan}{sgh,seri}
  \icmlauthor{Daniel Ting}{sgh,seri}
\end{icmlauthorlist}

\begin{center}
\small
\textsuperscript{1}Singapore Health Services, Singapore\
\textsuperscript{2}Singapore Eye Research Institute, Singapore\
\end{center}

\printAffiliationsAndNotice{}

\begin{abstract}
Transfer learning followed by fine-tuning is widely adopted in medical image classification due to consistent gains in diagnostic performance. However, in multi-class settings with overlapping visual features, improvements in accuracy do not guarantee stability of the visual evidence used to support predictions. We define \emph{semantic drift} as systematic changes in the attribution structure supporting a model’s predictions between transfer learning and full fine-tuning, reflecting potential shifts in underlying visual reasoning despite stable classification performance. Using a five-class chest X-ray task, we evaluate DenseNet201, ResNet50V2, and InceptionV3 under a two-stage training protocol and quantify drift with reference-free metrics capturing spatial localization and structural consistency of attribution maps. Across architectures, coarse anatomical localization remains stable, while overlap IoU reveals pronounced architecture-dependent reorganization of evidential structure. Extending beyond single-method analysis, stability rankings can reverse across LayerCAM and Grad-CAM++ under converged predictive performance, establishing explanation stability as an interaction between architecture, optimization phase, and attribution objective.
\end{abstract}

\vspace{0.2em}
]

\section{Introduction}
\label{sec:intro}

Transfer learning dominates contemporary medical image classification pipelines \citep{raghu2019transfusion,kolesnikov2020bit}, yet explanation stability across transfer learning and fine-tuning is rarely examined. Attribution maps can change substantially without accuracy degradation \citep{adebayo2018sanity,ghorbani2019interpretation,kindermans2019reliability}, and disagreement across explainers is pervasive in practice \citep{krishna2022disagreement,han2022nofree}. In safety-critical settings, this dissociation is consequential: models may achieve comparable predictive performance while relying on different visual evidence, yielding unstable clinical narratives even under correct predictions.

We define \textbf{semantic drift} as systematic changes in the visual evidence supporting a model’s predictions between transfer learning and full fine-tuning, reflecting potential shifts in internal visual reasoning while classification remains stable. Attribution behavior is method-contingent: gradient-based explainers optimize different computational objectives \citep{ancona2018understanding}, and stability measured under a single method is inherently scoped to that objective.

\textbf{Contributions.} This study provides three empirical contributions: (1) we quantify semantic drift during fine-tuning in a multi-class chest X-ray task using reference-free stability metrics capturing both spatial localization and structural consistency of attribution maps; (2) we restrict analysis to true-positive samples across training phases to isolate explanation evolution from changes in predictive correctness; and (3) we demonstrate that architecture-dependent stability rankings can reverse across LayerCAM and Grad-CAM++ even after predictive performance converges.

\section{Background}
\label{sec:background}

Gradient-based attribution methods are not interchangeable. \textbf{Grad-CAM} \citep{selvaraju2017gradcam} pools gradients to weight feature maps uniformly; \textbf{Grad-CAM++} \citep{chattopadhay2018gradcamplusplus} introduces higher-order gradients for adaptive weighting; \textbf{LayerCAM} \citep{jiang2021layercam} preserves fine-grained spatial detail via pixel-wise gradients. These methods encode different notions of importance \citep{ancona2018understanding}, and prior work documents sensitivity and disagreement in explanation behavior \citep{adebayo2018sanity,ghorbani2019interpretation,kindermans2019reliability,krishna2022disagreement,han2022nofree}. In chest X-ray interpretation, saliency methods also lag human localization benchmarks \citep{saporta2022benchmarking}, motivating stability analyses that do not conflate accuracy gains with explanation reliability.

\section{Methods}
\label{sec:method}

\subsection{Task, Architectures, and Training Protocol}

We conduct five-class chest X-ray classification (Normal, Pneumonia, Tuberculosis, COVID-19, Lung Opacity) on 11{,}733 training images, 1{,}675 validation images, and 3{,}354 test images. We evaluate three ImageNet-pretrained architectures: DenseNet201, ResNet50V2, and InceptionV3. Training follows a two-phase protocol: transfer learning with frozen backbones (epochs 1--10, Adam, learning rate $10^{-4}$), followed by full fine-tuning (epochs 11--20, learning rate $10^{-5}$). We compare epoch 8 (transfer-learning plateau) against epoch 19 (fine-tuning convergence) to maximize drift contrast while avoiding early training instability.

\begin{figure}[t]
\centering
\includegraphics[width=0.95\columnwidth]{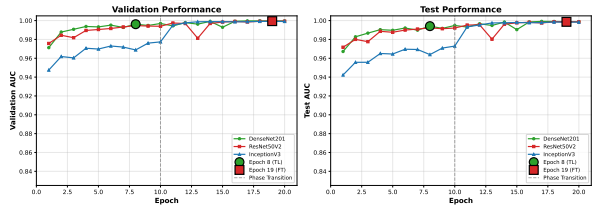}
\caption{\textbf{Epoch selection justification.} Epoch 8 marks the transfer learning plateau; epoch 19 represents fine-tuning convergence.}
\label{fig:epochs}
\end{figure}

\subsection{Attribution Methods}

We compute attribution maps using LayerCAM and Grad-CAM++ on penultimate convolutional layers. Maps are normalized to $[0,1]$ and thresholded at $\tau=0.2$ to isolate salient regions. Layer choices are: \texttt{conv5\_block32\_concat} (DenseNet201), \texttt{conv5\_block3\_out} (ResNet50V2), and \texttt{mixed10} (InceptionV3).

\subsection{True-Positive Filtering and Class Weighting}

To isolate explanation evolution from predictive correctness changes, a sample $(x_i,y_i)$ is included only if correctly classified at both epochs for all three architectures. Of 3{,}354 test samples, 2{,}430 (72.5\%) satisfy this criterion. To address class imbalance, semantic drift metrics are aggregated using inverse-frequency class weighting (Table~\ref{tab:dataset}).

\begin{table}[t]
\centering
\caption{Dataset composition and inverse-frequency weights ensuring proportional class contributions despite imbalance.}
\label{tab:dataset}
\begin{small}
\setlength{\tabcolsep}{6pt}
\begin{tabular}{lrrr}
\toprule
\textbf{Class} & \textbf{Test Samples} & \textbf{\%} & \textbf{Weight} \\
\midrule
Normal & 317 & 9.5 & 0.235 \\
Pneumonia & 855 & 25.5 & 0.087 \\
Tuberculosis & 141 & 4.2 & 0.528 \\
COVID-19 & 839 & 25.0 & 0.089 \\
Lung Opacity & 1202 & 35.8 & 0.062 \\
\midrule
\textbf{Total} & \textbf{3354} & \textbf{100.0} & \textbf{1.000} \\
\bottomrule
\end{tabular}
\end{small}
\end{table}

\subsection{Semantic Drift Metrics}

We quantify semantic drift using reference-free metrics capturing spatial localization and structural consistency.

\textbf{Spatial displacement} measures normalized center-of-mass movement:
\begin{equation}
\Delta_{\text{spatial}} =
\frac{\left\|\mathrm{CoM}(\tilde{A}_{\text{TL}})-\mathrm{CoM}(\tilde{A}_{\text{FT}})\right\|_2}{\sqrt{h^2+w^2}}.
\end{equation}

\textbf{Overlap IoU} (primary drift metric) measures preservation of discriminative structure:
\begin{equation}
\mathrm{IoU}=\frac{|M_{\text{TL}}\cap M_{\text{FT}}|}{|M_{\text{TL}}\cup M_{\text{FT}}|},\quad
M=\mathbb{1}[\tilde{A}>\tau].
\end{equation}

We additionally report \textbf{pattern correlation} (Pearson correlation between continuous maps) and \textbf{concentration change} (Shannon entropy difference) to characterize continuous similarity and attention redistribution.

\subsection{Formalizing Semantic Drift as Cross-Phase Evidence Transformation}

Let $f_{\theta^{\text{TL}}}$ denote the model after transfer learning and
$f_{\theta^{\text{FT}}}$ denote the model after full fine-tuning.
For an input image $x$, let $\mathcal{A}(x; f_\theta, \phi)$ denote the attribution map
produced by explainer $\phi$ (e.g., LayerCAM, Grad-CAM++).

Semantic drift can be viewed as the transformation:
\begin{equation}
\mathcal{D}(x; \phi) =
\Delta \big( \mathcal{A}(x; f_{\theta^{\text{TL}}}, \phi),
              \mathcal{A}(x; f_{\theta^{\text{FT}}}, \phi) \big)
\end{equation}
where $\Delta(\cdot,\cdot)$ denotes a stability operator. In this study,
$\Delta$ includes spatial displacement, overlap IoU, pattern correlation,
and entropy-based concentration change.

Importantly, $\mathcal{D}(x; \phi)$ is conditioned on:
(i) architecture,
(ii) optimization phase,
and (iii) attribution objective.
Thus, explanation stability is not solely a property of $f_\theta$,
but of the triplet $(f_\theta, \phi, \text{training phase})$.

We aggregate $\mathcal{D}(x; \phi)$ across samples using inverse-frequency
class weighting to ensure that rare pathologies contribute proportionally
to overall stability estimates. This prevents dominant classes (e.g.,
Lung Opacity) from masking architecture-dependent instability in minority classes.

\section{Results}
\label{sec:results}

\subsection{Predictive Performance Converges After Fine-Tuning}

All architectures achieve high predictive performance at epoch 19 (Table~\ref{tab:performance}), demonstrating comparable classification capability despite divergent explanation behavior.

\begin{table}[t]
\centering
\caption{Test performance at epoch 19 (fine-tuned). All architectures achieve $>$99\% AUC with comparable accuracy and F1 scores, demonstrating equivalent predictive capability despite divergent explanation stability.}
\label{tab:performance}
\begin{small}
\begin{tabular}{lccc}
\toprule
\textbf{Architecture} & \textbf{AUC} & \textbf{Accuracy} & \textbf{F1-Score} \\
\midrule
DenseNet201 & 0.995 & 0.936 & 0.935 \\
ResNet50V2  & 0.998 & 0.973 & 0.959 \\
InceptionV3 & 0.998 & 0.973 & 0.964 \\
\bottomrule
\end{tabular}
\end{small}
\end{table}

\subsection{Architecture-Dependent Semantic Drift Under LayerCAM}

Under LayerCAM, spatial displacement remains low and tightly bounded (Table~\ref{tab:method_summary}), indicating preserved coarse anatomical localization during fine-tuning. In contrast, overlap IoU reveals architecture-dependent differences in structural consistency: InceptionV3 achieves the highest overlap (0.777$\pm$0.128), followed by DenseNet201 (0.699$\pm$0.171), while ResNet50V2 exhibits lower overlap (0.519$\pm$0.154). These results show that preserved spatial localization can mask substantial reorganization of evidential structure, and that spatial alignment alone is insufficient to characterize explanation stability.

\subsection{Cross-Method Evaluation Reveals Ranking Reversal}

\begin{figure*}[t]
\centering
\includegraphics[width=0.82\textwidth]{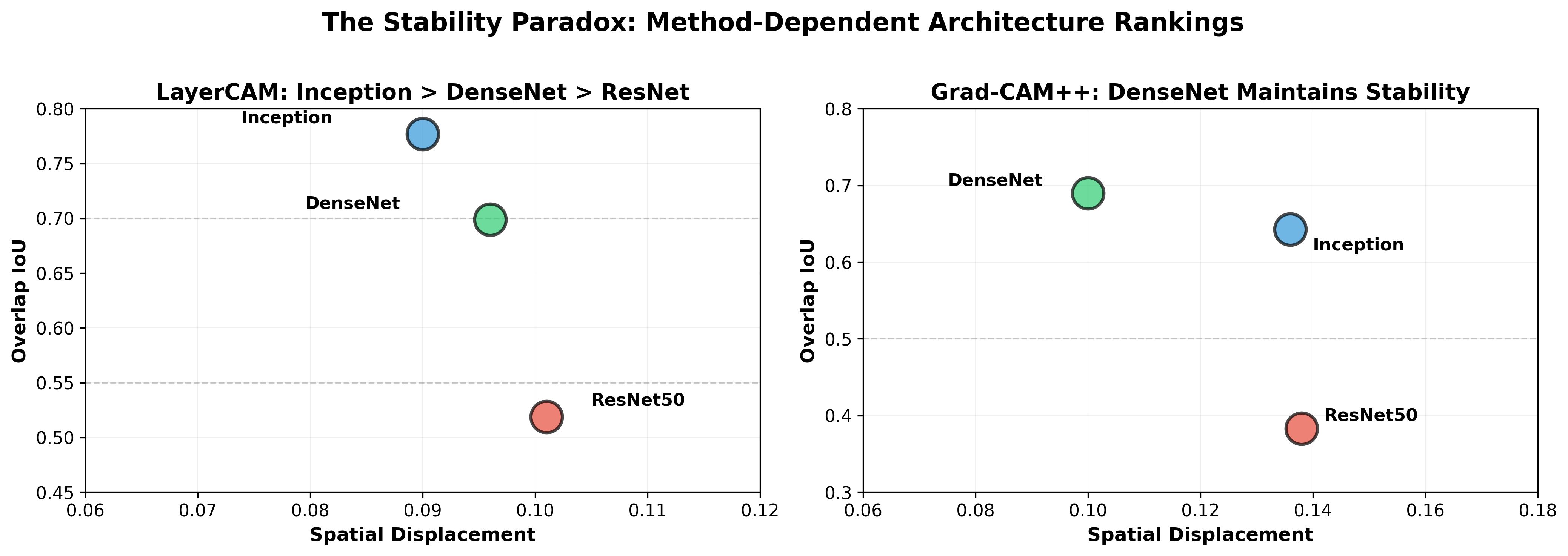}
\caption{\textbf{Method-dependent stability rankings reveal architectural reversal.} \textbf{Left (LayerCAM):} InceptionV3 (IoU=0.777) outperforms DenseNet201 (0.699) and ResNet50V2 (0.519). \textbf{Right (Grad-CAM++):} DenseNet201 (0.690) leads InceptionV3 (0.643) and ResNet50V2 (0.383).}
\label{fig:paradox}
\end{figure*}

Figure~\ref{fig:paradox} and Table~\ref{tab:method_summary} show that stability rankings can reverse across attribution objectives despite converged predictive performance. Under Grad-CAM++, DenseNet201 becomes the most stable architecture (IoU 0.690$\pm$0.169), while InceptionV3 decreases to 0.643$\pm$0.172 and ResNet50V2 collapses to 0.383$\pm$0.174. DenseNet201 exhibits minimal cross-method variation (0.699$\rightarrow$0.690), whereas InceptionV3 shows pronounced method dependency (0.777$\rightarrow$0.643). This establishes method sensitivity as a measurable dimension of explanation robustness.

\subsection{Distributional Characteristics of Drift}

Beyond mean stability values, variance patterns reveal additional architecture-specific behavior. ResNet50V2 demonstrates both lower mean overlap IoU and higher inter-sample variance under Grad-CAM++, suggesting heterogeneous internal reorganization across cases. In contrast, DenseNet201 exhibits comparatively narrow variance across both attribution methods, indicating more uniform refinement of evidential structure during fine-tuning.

The dissociation between spatial displacement and overlap IoU is particularly notable. Across architectures, spatial displacement remains tightly bounded ($\le 0.14$), indicating preserved coarse anatomical focus. However, overlap IoU ranges from 0.383 to 0.777 depending on architecture and explainer. This confirms that center-of-mass alignment alone fails to capture structural reconfiguration of discriminative regions.

Pattern correlation and concentration change further clarify these differences. For example, ResNet50V2 under Grad-CAM++ shows pronounced negative concentration change ($-0.516 \pm 0.516$), indicating redistribution of attention mass, whereas DenseNet201 maintains near-zero concentration change across both methods. These results suggest that dense connectivity may regularize feature reuse during fine-tuning, leading to more coherent explanation evolution.

\begin{table*}[t]
\centering
\caption{Method-dependent semantic drift metrics (weighted, $N=2430$ true-positive test images).}
\label{tab:method_summary}
\begin{small}
\begin{tabular}{llcccc}
\toprule
\textbf{Method} & \textbf{Architecture} & \textbf{Spatial Disp} & \textbf{Overlap IoU} & \textbf{Pattern Corr} & \textbf{Conc Change} \\
\midrule
\multirow{3}{*}{LayerCAM}
& DenseNet201 & 0.096 $\pm$ 0.074 & 0.699 $\pm$ 0.171 & 0.368 $\pm$ 0.337 & $-$0.050 $\pm$ 0.136 \\
& ResNet50V2  & 0.101 $\pm$ 0.062 & 0.519 $\pm$ 0.154 & 0.403 $\pm$ 0.285 & $-$0.136 $\pm$ 0.130 \\
& InceptionV3 & 0.090 $\pm$ 0.058 & \textbf{0.777} $\pm$ 0.128 & 0.220 $\pm$ 0.465 & $-$0.024 $\pm$ 0.077 \\
\midrule
\multirow{3}{*}{Grad-CAM++}
& DenseNet201 & 0.100 $\pm$ 0.073 & \textbf{0.690} $\pm$ 0.169 & 0.345 $\pm$ 0.350 & $-$0.049 $\pm$ 0.172 \\
& ResNet50V2  & 0.138 $\pm$ 0.085 & 0.383 $\pm$ 0.174 & 0.506 $\pm$ 0.246 & $-$0.516 $\pm$ 0.516 \\
& InceptionV3 & 0.136 $\pm$ 0.073 & 0.643 $\pm$ 0.172 & 0.386 $\pm$ 0.423 & $+$0.275 $\pm$ 0.303 \\
\bottomrule
\end{tabular}
\end{small}
\end{table*}

\subsection{Qualitative Evidence of Method-Robust Stability in DenseNet201}

\begin{figure*}[t]
\centering
\includegraphics[width=0.92\textwidth]{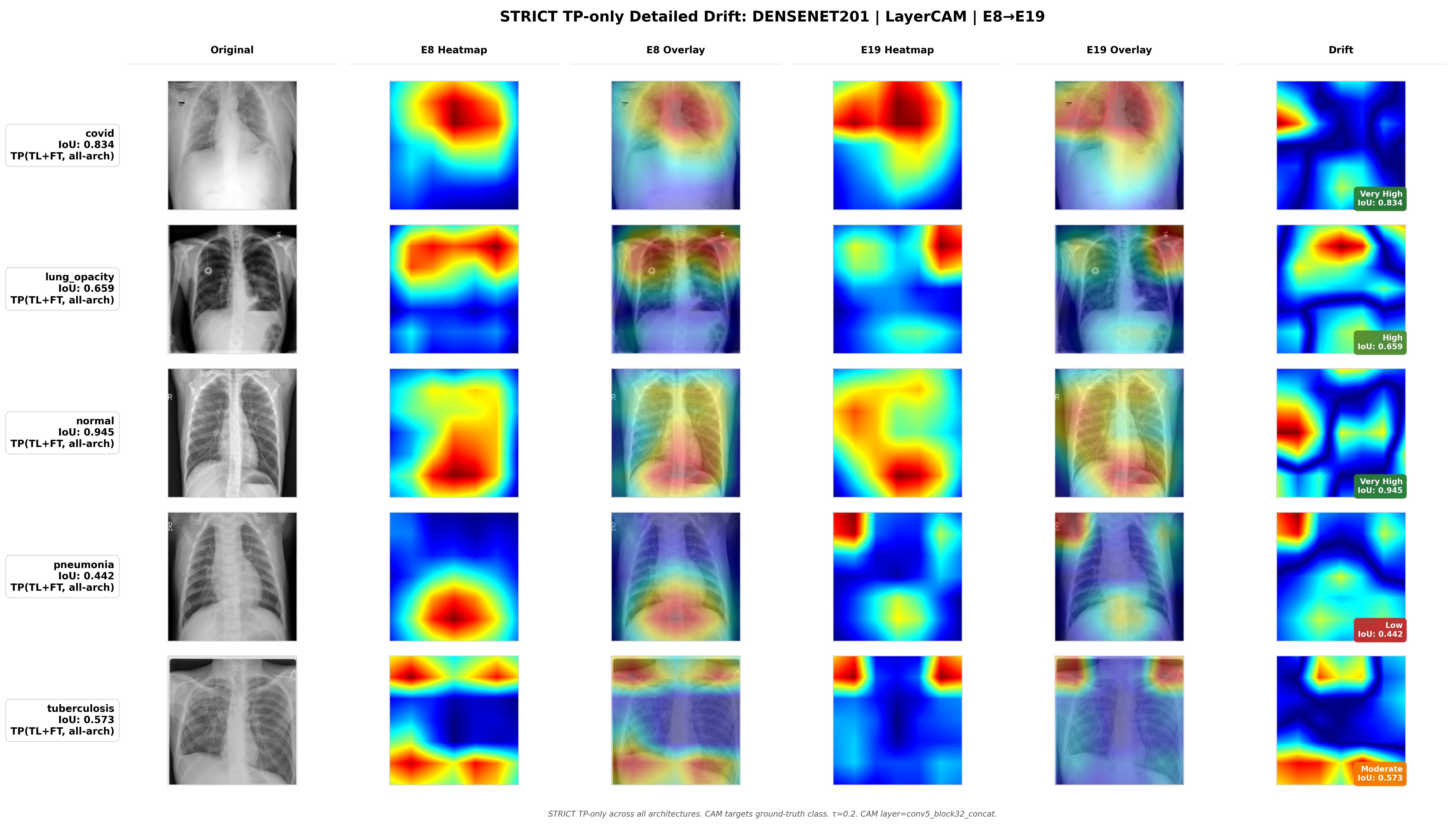}
\vspace{0.35cm}
\includegraphics[width=0.92\textwidth]{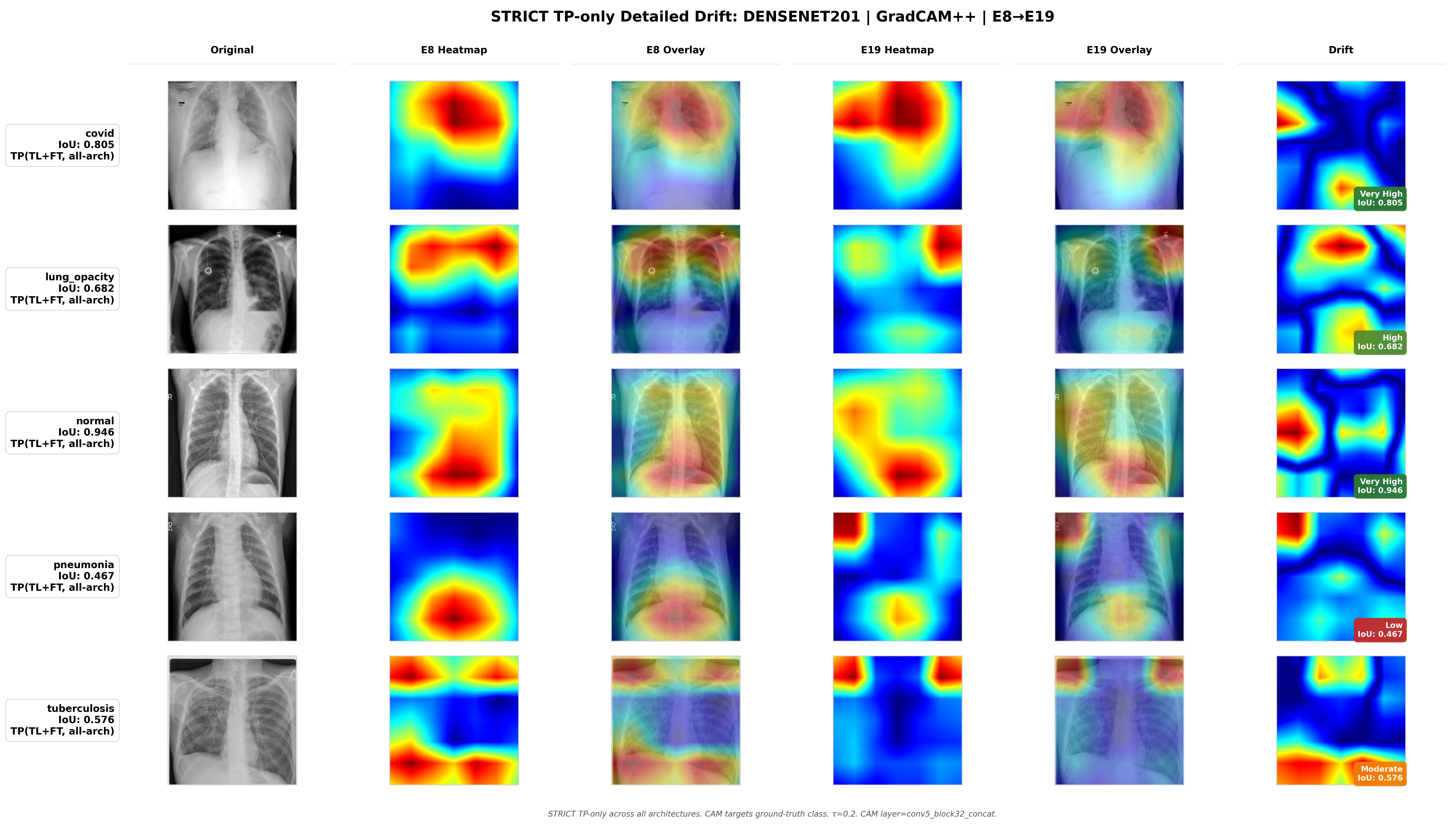}
\caption{\textbf{DenseNet201 cross-method comparison: LayerCAM (top) vs Grad-CAM++ (bottom).} Dense connectivity yields coherent explanation refinement across training phases with limited cross-method divergence in salient structure.}
\label{fig:densenet_compare}
\end{figure*}

To contextualize method-robust behavior, we include DenseNet201 qualitative overlays under LayerCAM and Grad-CAM++ (Figure~\ref{fig:densenet_compare}). Dense connectivity yields coherent refinement across training phases with limited cross-method divergence in salient structure.

\section{Discussion}
\label{sec:discussion}

\subsection{What the Drift Metrics Reveal}

Across architectures, semantic drift exhibits a consistent pattern: coarse anatomical localization can remain stable while the \emph{structure} of discriminative evidence reorganizes. This is reflected by low spatial displacement alongside architecture- and method-dependent overlap IoU. In multi-class settings with overlapping radiographic signatures, these shifts can materially alter the narrative a clinician would infer from visual explanations, even when predictions remain correct.

\subsection{Applicability}

This work is practically useful in three ways.

\textbf{Post-fine-tuning explanation auditing.} Fine-tuning is routinely performed to improve downstream performance, but explanations are rarely audited across optimization phases. Semantic drift provides a reference-free way to quantify whether evidence patterns remain coherent after fine-tuning without requiring pixel-level ground truth.

\textbf{Architecture selection when accuracy converges.} When multiple backbones achieve near-identical predictive metrics, semantic drift offers an additional reliability axis: the degree to which the evidential structure remains consistent and robust to attribution objective.

\textbf{A reference-free building block for evaluation frameworks.} Reference-based localization benchmarks are valuable but costly and incomplete \citep{saporta2022benchmarking}. Drift metrics operate without pixel-level ground truth and can be applied across datasets, pathologies, and training regimes. Toolkits emphasize multi-metric evaluation \citep{hedstrom2023quantus,hedstrom2023metaquantus}; drift metrics complement these efforts by capturing cross-phase stability (transfer learning $\rightarrow$ fine-tuning) and cross-method sensitivity (LayerCAM $\leftrightarrow$ Grad-CAM++). In future reference-free frameworks, drift can serve as an “evidence continuity” component: models that improve accuracy but substantially change evidential structure can be flagged for review even when conventional performance monitoring would remain silent.

\subsection{Implications for Explanation-Aware Model Development}

The observed semantic drift has implications beyond descriptive stability measurement. In medical imaging pipelines, fine-tuning is routinely performed to improve domain adaptation performance. However, performance monitoring typically focuses on predictive metrics alone. Our findings demonstrate that fine-tuning can preserve accuracy while reorganizing evidential structure in an architecture- and method-dependent manner.

This suggests three practical extensions.

\textbf{(1) Cross-Phase Stability Auditing.}
Explanation stability can be evaluated at predefined checkpoints (e.g., post-transfer learning vs. post-fine-tuning) to detect silent evidence shifts. Because drift metrics are reference-free, this procedure can be implemented without pixel-level annotations, making it scalable across datasets and institutions.

\textbf{(2) Architecture Selection Under Saturated Performance.}
When multiple backbones achieve near-identical AUC and F1, semantic drift provides an orthogonal reliability dimension. Architectures with minimal cross-phase and cross-method variation may offer more coherent internal evidence evolution, which is relevant for deployment in high-stakes environments.

\textbf{(3) Integration into Reference-Free Evaluation Frameworks.}
Existing explainability benchmarks often rely on external annotations or perturbation-based faithfulness metrics. Semantic drift complements these approaches by quantifying \emph{evidence continuity} across optimization stages. In future evaluation frameworks, stability across training phases and attribution objectives could serve as a structural robustness criterion alongside accuracy and calibration.

\subsection{Limitations and Scope}

Several limitations should be noted. First, this study evaluates two gradient-based attribution methods and three convolutional architectures. Transformer-based models and perturbation-based explainers may exhibit different drift dynamics. Second, drift metrics quantify consistency rather than correctness; high stability does not guarantee alignment with clinician-defined evidence. Third, analysis is restricted to true-positive cases to isolate explanation evolution. While this controls for prediction changes, it does not capture drift behavior in decision-boundary samples.

Finally, semantic drift is evaluated between two discrete training checkpoints. Continuous tracking across all epochs may reveal more nuanced temporal patterns of evidence evolution.

\section{Conclusion}
\label{sec:conclusion}

We quantified semantic drift during fine-tuning in a five-class chest X-ray task using reference-free stability metrics. Across architectures, coarse localization remains stable, while overlap IoU reveals architecture-dependent reorganization of evidential structure. Extending beyond single-method evaluation, stability rankings reverse between LayerCAM and Grad-CAM++ despite converged predictive performance, establishing method sensitivity as a measurable aspect of explanation robustness. These results support explanation-aware reporting that separates localization stability from structural consistency and characterizes sensitivity to attribution method choice.

\bibliographystyle{unsrtnat}
\bibliography{references}

@inproceedings{raghu2019transfusion,
  title     = {Transfusion: Understanding Transfer Learning for Medical Imaging},
  author    = {Raghu, Maithra and Zhang, Chiyuan and Kleinberg, Jon and Bengio, Samy},
  booktitle = {Advances in Neural Information Processing Systems (NeurIPS)},
  volume    = {32},
  pages     = {3342--3352},
  year      = {2019}
}

@inproceedings{kolesnikov2020bit,
  title     = {Big Transfer ({BiT}): General Visual Representation Learning},
  author    = {Kolesnikov, Alexander and Beyer, Lucas and Zhai, Xiaohua and Puigcerver, Joan and Yung, Jessica and Gelly, Sylvain and Houlsby, Neil},
  booktitle = {European Conference on Computer Vision (ECCV)},
  pages     = {491--507},
  year      = {2020},
  organization = {Springer}
}

@inproceedings{selvaraju2017gradcam,
  title     = {{Grad-CAM}: Visual Explanations from Deep Networks via Gradient-Based Localization},
  author    = {Selvaraju, Ramprasaath R and Cogswell, Michael and Das, Abhishek and Vedantam, Ramakrishna and Parikh, Devi and Batra, Dhruv},
  booktitle = {Proceedings of the IEEE International Conference on Computer Vision (ICCV)},
  pages     = {618--626},
  year      = {2017}
}

@inproceedings{chattopadhay2018gradcamplusplus,
  title     = {{Grad-CAM++}: Generalized Gradient-Based Visual Explanations for Deep Convolutional Networks},
  author    = {Chattopadhay, Aditya and Sarkar, Anirban and Howlader, Prantik and Balasubramanian, Vineeth N},
  booktitle = {2018 IEEE Winter Conference on Applications of Computer Vision (WACV)},
  pages     = {839--847},
  year      = {2018},
  organization = {IEEE}
}

@article{jiang2021layercam,
  title     = {{LayerCAM}: Exploring Hierarchical Class Activation Maps for Localization},
  author    = {Jiang, Peng-Tao and Zhang, Chang-Bin and Hou, Qibin and Cheng, Ming-Ming and Wei, Yunchao},
  journal   = {IEEE Transactions on Image Processing},
  volume    = {30},
  pages     = {5875--5888},
  year      = {2021},
  publisher = {IEEE}
}

@inproceedings{ancona2018understanding,
  title     = {Towards Better Understanding of Gradient-Based Attribution Methods for Deep Neural Networks},
  author    = {Ancona, Marco and Ceolini, Enea and \"Oztireli, Cengiz and Gross, Markus},
  booktitle = {International Conference on Learning Representations (ICLR)},
  year      = {2018}
}

@inproceedings{adebayo2018sanity,
  title     = {Sanity Checks for Saliency Maps},
  author    = {Adebayo, Julius and Gilmer, Justin and Muelly, Michael and Goodfellow, Ian and Hardt, Moritz and Kim, Been},
  booktitle = {Advances in Neural Information Processing Systems (NeurIPS)},
  volume    = {31},
  pages     = {9505--9515},
  year      = {2018}
}

@inproceedings{ghorbani2019interpretation,
  title     = {Interpretation of Neural Networks is Fragile},
  author    = {Ghorbani, Amirata and Abid, Abubakar and Zou, James},
  booktitle = {Proceedings of the AAAI Conference on Artificial Intelligence},
  volume    = {33},
  number    = {01},
  pages     = {3681--3688},
  year      = {2019}
}

@incollection{kindermans2019reliability,
  title     = {The (Un)reliability of Saliency Methods},
  author    = {Kindermans, Pieter-Jan and Hooker, Sara and Adebayo, Julius and Alber, Maximilian and Sch{\"u}tt, Kristof T and D{\"a}hne, Sven and Erhan, Dumitru and Kim, Been},
  booktitle = {Explainable AI: Interpreting, Explaining and Visualizing Deep Learning},
  pages     = {267--280},
  year      = {2019},
  publisher = {Springer}
}

@article{krishna2022disagreement,
  title   = {The Disagreement Problem in Explainable Machine Learning: A Practitioner's Perspective},
  author  = {Krishna, Satyapriya and Han, Tessa and Gu, Alex and Pombra, Javin and Jabbari, Shahin and Wu, Steven and Lakkaraju, Himabindu},
  journal = {Transactions on Machine Learning Research},
  year    = {2024},
  note    = {arXiv:2202.01602 (2022); TMLR publication (2024)}
}

@inproceedings{han2022nofree,
  title     = {Which Explanation Should I Choose? A Function Approximation Perspective to Characterizing Post Hoc Explanations},
  author    = {Han, Tessa and Srinivas, Suraj and Lakkaraju, Himabindu},
  booktitle = {Advances in Neural Information Processing Systems (NeurIPS)},
  volume    = {35},
  pages     = {22121--22132},
  year      = {2022},
  note      = {arXiv:2206.01254}
}

@article{saporta2022benchmarking,
  title     = {Benchmarking Saliency Methods for Chest X-ray Interpretation},
  author    = {Saporta, Adriel and Gui, Xiaotong and Agrawal, Ashwin and Pareek, Anuj and Truong, Steven QH and Nguyen, Chanh DT and Ngo, Van-Doan and Seekins, Jayne and Blankenberg, Francis G and Ng, Andrew Y and Lungren, Matthew P and Rajpurkar, Pranav},
  journal   = {Nature Machine Intelligence},
  volume    = {4},
  number    = {10},
  pages     = {867--878},
  year      = {2022},
  publisher = {Nature Publishing Group}
}

@article{hedstrom2023quantus,
  title   = {Quantus: An Explainable AI Toolkit for Responsible Evaluation of Neural Network Explanations and Beyond},
  author  = {Hedstr{\"o}m, Anna and Weber, Leander and Krakowczyk, Daniel and Bareeva, Dilyara and Motzkus, Franz and Samek, Wojciech and Lapuschkin, Sebastian and M{\"u}ller, Marina M-C},
  journal = {Journal of Machine Learning Research},
  volume  = {24},
  number  = {34},
  pages   = {1--11},
  year    = {2023}
}

@article{hedstrom2023metaquantus,
  title   = {The Meta-Evaluation Problem in Explainable AI: Identifying Reliable Estimators with MetaQuantus},
  author  = {Hedstr{\"o}m, Anna and Bommer, Philine and Wickstr{\"o}m, Kristoffer K and Samek, Wojciech and Lapuschkin, Sebastian and M{\"u}ller, Marina M-C},
  journal = {Transactions on Machine Learning Research},
  year    = {2023},
  note    = {Featured Certification}
}

\end{document}